\documentclass[letterpaper,twocolumn,fleqn]{article} 
\usepackage{ist}
\usepackage{stackengine}
\usepackage{amsmath}
\usepackage{amssymb}
\usepackage{hyperref}
\newcommand{\norm}[1]{\left\lVert#1\right\rVert}
\title{Virtual Adversarial Training in Feature Space to Improve Unsupervised Video Domain Adaptation}
\author{Artjoms Gorpincenko, Geoffrey French, Michal Mackiewicz. \\ School of Computing Sciences, University of East Anglia, Norwich, England.}
\date{}
\begin{document} 
\maketitle 
\thispagestyle{empty}

\begin{abstract}
Virtual Adversarial Training has recently seen a lot of success in semi-supervised learning, as well as unsupervised Domain Adaptation. However, so far it has been used on input samples in the pixel space, whereas we propose to apply it directly to feature vectors. We also discuss the unstable behaviour of entropy minimization and Decision-Boundary Iterative Refinement Training With a Teacher in Domain Adaptation, and suggest substitutes that achieve similar behaviour. By adding the aforementioned techniques to the state of the art model TA$^3$N, we either maintain competitive results or outperform prior art in multiple unsupervised video Domain Adaptation tasks. 
\end{abstract}

\section{Introduction}\label{Introduction}
Deep convolutional neural networks (CNNs) have become the standard approach for a variety of machine learning tasks, due to their strong performance and ability to learn the most useful features directly from the data. However, when the amount of labeled training data is limited or unavailable, the performance of deep feed-forward models drops significantly. Acquiring and labeling new imagery for the task of interest is often expensive, tedious, or even impossible. Ideally, a classifier trained using a labeled dataset could be applied to a related target dataset, achieving the same task \cite{virtual}. This approach frequently comes at the cost of domain shift - a scenario where the training data distribution (the source domain) is different to the test data distribution (the target domain). In computer vision, this normally manifests as differences in appearance that arise due to different lighting or image capture conditions. CNNs are very sensitive to such shifts and generally fail to perform well, even when the gap between visual domains is small \cite{predictive}. Learning a model in the presence of a shift between labeled source data and unlabeled target data is an instance of Unsupervised Domain Adaptation (UDA), the most challenging setting of Domain Adaptation (DA). Although recently there have been many advancements in image-based UDA \cite{dann, se, dirt-t, vmt}, video-based UDA remains an under-explored research topic.

Data augmentation has seen a lot of success in semi-supervised learning, thanks to the ability of enlarging training distribution through virtually created samples \cite{mixup, cutmix, vatpaper, fixmatch}. Due to the similar nature of the task, this powerful technique was also used in Domain Adaptation, and delivered state of the art results \cite{se, dirt-t, vmt}. However, so far the methods considered applying augmentation to pixel space only, not affecting the feature space directly. In this paper, we explore the benefits of Virtual Adversarial Training (VAT) \cite{vatpaper} in feature space, and apply it to the video-based UDA technique called Temporal Attentive Adversarial Adaptation Network (TA$^3$N) \cite{ta3npaper} to show its effectiveness. Together with other proposed changes, we either maintain competitive results or achieve state of the art performance on 6 distinct adaptation paths, formed by 3 publicly available video datasets. 

\section{Background}\label{Background}
\subsection{Domain Adaptation}
The main objective of most DA approaches is to find a common feature space between the source and target domains through an optimization process utilizing multiple loss functions. This idea often results in an architecture that follows two paths, one for each domain. Traditionally, the source path is accompanied by classification loss, whereas domain loss drives source and target feature distributions closer to each other. Our work improves on an adversarial-based DA method, an instance where a domain Discriminator network is utilized to classify two sets of features, while the feature extractor uses the gradient reversal layer (GRL) \cite{dann}, to allow for simultaneous backpropagation. Many modern DA methods use the non-saturating Generative Adversarial Network (GAN) loss \cite{gan} and inverted labels instead, and  split the optimization process into two parts \cite{dirt-t, vmt}, but the underlying idea remains the same.

\subsection{Video Domain Adaptation}
Unlike static images, videos are comprised of moving visual frames, which may introduce domain shifts in the temporal dimension, as well as the spatial one. We hypothesize that the main reason as to why this issue has not been explicitly addressed by most approaches so far, is because they focused on small-scale video-based DA datasets, where the domain gap along the temporal direction is rather small \cite{sultani, xu, jamal}. Chen et al. \cite{ta3npaper} solves this by introducing two new large-scale datasets, in addition to proposing TA$^3$N, that encodes the relation between frames into features and aligns local temporal dynamics with larger domain discrepancy through the domain attention mechanism. Although the aim of our work is not to improve on the temporal encoding or feature extraction itself, we still extends TA$^3$N by showing that Virtual Adversarial Training can be applied in feature space to help solve video-based UDA problems.

\subsection{Virtual Adversarial Training}
Virtual Adversarial Training is a regularization method that was initially introduced for supervised and semi-supervised learning \cite{vatpaper}. VAT incorporates the locally-Lipschitz constraint to the given training data points via local perturbation and is able to define the adversarial direction without label information. This results in a smooth and robust model, which is otherwise difficult to achieve when semi-supervised and unsupervised tasks are considered, due to the lack of labels. VAT was also applied to image-based UDA \cite{dirt-t, vmt}, and achieved state of the art results. However, the aforementioned methods used VAT in pixel space, whereas the idea of expanding clusters of data points can also be applied to feature space, which allows for faster computations and does not interfere with the feature extraction part of a system, that often might work as a separate model in video classification \cite{ta3npaper, piergiovanni, mao}.

\section{Method}\label{Method}
We begin by briefly describing the TA$^3$N model and losses used for training the network. First, raw videos $V$ are passed through the ResNet-101 model $G_{r}$, that was initially pre-trained on ImageNet, to produce general-purpose features $\hat X$, for each frame. Then, the vectors are converted to task-driven features $\hat t$, by the Spatial module $G_{sf}$, where the task is video classification. Next, the Temporal module $G_{tf}$, encodes the relation between frames into single vectors, by passing time-ordered sets of 5 frame representations through a multilayer perceptron (MLP), and sums them together, resulting in one feature vector $\hat d$, for each video. Lastly, the video features go through a fully-connected layer $G_y$, which produces the final predictions $\hat y$. The system makes sure that source and target domains are aligned at all stages by having domain Discriminators included in Spatial, Relation and Temporal modules:
\begin{equation}
 L_{d} = - \frac{1}{N_{S\cup T}} \sum_{i=1}^{N_{S\cup T}} (\lambda^s L_{sd}^i + \lambda^r L_{rd}^i + \lambda^t L_{td}^i)
\end{equation}
where $N_{S\cup T}$ is the number of training samples in both source ($S$) and target ($T$) domains, $L_{*}^i$ is calculated via the cross entropy loss function which is applied to Discriminators' predictions and domain labels, and $\lambda$ parameters control the weighting of each module. The authors also minimize the entropy for samples that have low domain discrepancy, via the attentive entropy loss:
\begin{equation}
L_{ae} = \frac{1}{N_{S\cup T}} \sum_{i=1}^{N_{S\cup T}} \lambda^{ae} (1 + H(\hat d_i)) \cdot H(\hat y_i)
\end{equation}
with $H(p)$ being the entropy function. Lastly, the traditional classification loss $L_{cls}$, is utilized to minimise the source domain error via cross entropy loss on supervised samples from the source domain.
Putting everything together results in the overall loss:
\begin{equation}
L_{TA^3N} = L_{d} + L_{ae} + L_{cls}
\end{equation}
Note that we omit many details to focus only on the relevant parts and save space. The full description is available in the TA$^3$N paper \cite{ta3npaper}.

\subsection{Virtual Adversarial Training in Feature Space}
For simplicity, let all the aforementioned modules together with the classification layer be $G_t$, such that
\begin{equation}
G_t(\hat X_i) = G_y(G_{tf}(G_{sf}(\hat X_i)))
\end{equation}
Traditionally, VAT would seek to minimize the following objective:
\begin{equation}
L_{vat} = \frac{1}{N_{S\cup T}} \sum_{i=1}^{N_{S\cup T}} \lambda^{vat} LDS(V_i)
\end{equation}
where
\begin{equation}
LDS(V_i) = \stackunder{max}{$\norm{r}\leq \epsilon$} D_{KL}(G_t(G_{r}(V_i)) \parallel G_t(G_{r}(V_i + r)))
\end{equation}
with $D_{KL}$ and $r$ being the Kullback–Leibler divergence and adversarial noise, respectively.
Following TA$^3$N, we freeze the weights of the ImageNet pre-trained ResNet-101, hence, using the adversarial perturbation in input pixel space $r_V$ will have little benefit in comparison to using the feature space perturbation $r_{\hat X}$. Given that adversarial perturbations are computed via backpropagation, $r_{\hat X}$ is an intermediate value used to compute $r_V$. Applying $r_V$ would most likely induce the ResNet model to produce features with the similar feature space perturbation to $r_{\hat X}$, which would then be used to train the subsequent layers in the network. Therefore, although adding noise to video frames seems logical, there is little benefit in expending the additional computation required to propagate back $r_{\hat X}$ through $G_r$ to obtain $r_V$, only to pass it forward through the ResNet again to recover $r_{\hat X}$. 

Perhaps, the biggest disadvantage of applying any kind of augmentation to feature vectors is the fact that unlike imagery, it is impossible to visually evaluate the effect of such operations. This introduces additional challenge when it comes to hyperparameter selection. In case of VAT, the main parameter is $\epsilon$, the norm constraint that determines the scale to which adversarial direction is applied to real samples. Such variables are often found via performing multiple experiments and choosing the one that gives the best result. However, this approach is inaccessible in real-world UDA scenarios, where the domain of interest is unlabeled. Moreover, input images usually lay in a predetermined pixel range, whereas feature vector values depend on several factors, such as losses, as well as normalization and regularization terms used within the network. These factors affect model weights through backpropagation and make the feature space more sensitive to perturbations than the pixel space. Miyato et al. \cite{vatpaper} note that for small $\epsilon$, both hyperparameters $\lambda^{vat}$ and $\epsilon$ have virtually the same effect on the strength of regularization, and therefore, tuning just one of them is sufficient. We follow their advice and fix $\epsilon = 1$. Therefore, the
local distributional smoothness (LDS) becomes:
\begin{equation}
LDS(\hat X_i) = \stackunder{max}{$\norm{r}\leq 1$} D_{KL}(G_t(\hat X_i) \parallel G_t(\hat X_i + r))
\end{equation}
Although the $\lambda^{vat}$ value could be found via extensive grid search, we noted that $\lambda^{vat} = 0.01$ consistently achieved good results, hence, we fixed it too, for simplicity. Therefore, our proposed loss becomes:
\begin{equation}
    L = L_{TA^3N} + \frac{1}{N_{S\cup T}} \sum_{i=1}^{N_{S\cup T}} \lambda^{vat} LDS(\hat X_i)
\end{equation}
We stress that Virtual Adversarial Training is applied to both source and target training features. 

\subsection{Entropy Minimization}
\begin{figure}[!h]
  \includegraphics[width=\columnwidth]{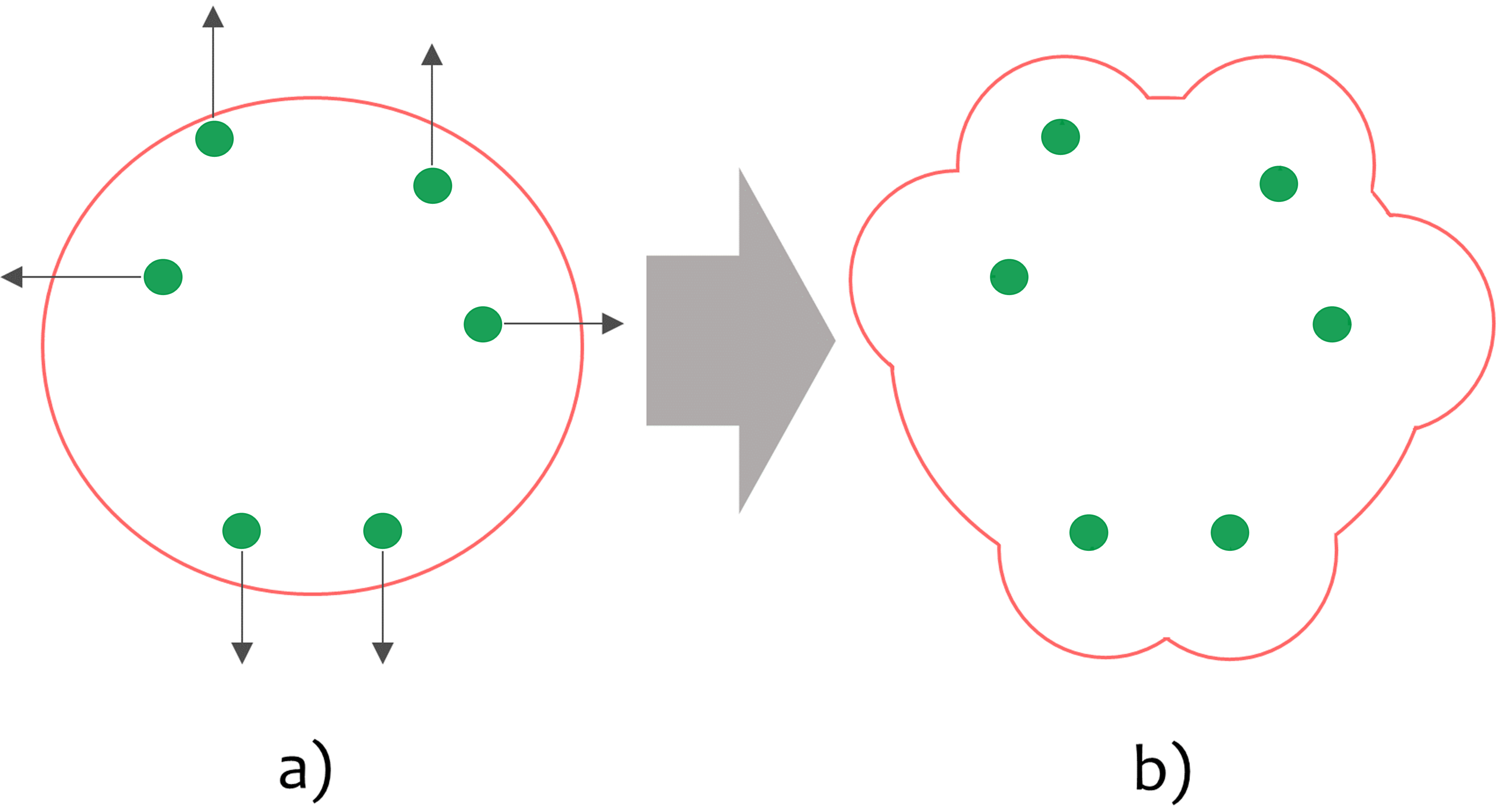}
  \caption{VAT cluster expansion. Green - input data points, red - classifier decision boundary, black - adversarial direction. \texttt{a)} new samples are created by prioritizing areas where the model has the highest change in predictions, i.e. outside or close to the decision boundary. \texttt{b)} by forcing the network to be consistent with its predictions, the cluster gets naturally expanded, indirectly making input data closer to its center.}
  \label{VAT}
\end{figure}

Cluster assumption for Domain Adaptation was extensively discussed in DIRT-T \cite{dirt-t}, where it was enforced via conditional entropy (CE) minimization. However, for it to be successful, the classifier needs to be locally-Lipschitz, which guarantees that the model predictions are smooth around data points. This constraint was then successfully accomplished by employing VAT. We hypothesize that similar behaviour can be achieved by using VAT only, as training the network to be locally consistent with its predictions naturally leads to cluster expansion, which indirectly makes the classifier confident about data that lay in the middle of those virtually created clusters (Fig. \ref{VAT}). Clearly, using CE and VAT together can yield better results than VAT or CE alone in the semi-supervised setting \cite{vatpaper, avital}. However, entropy minimization simply ensures that the model is extremely confident about each data point, instead of placing decision boundaries in low-density areas \cite{avital}. As the number of labeled samples decreases, resulting in little information about true class distributions, the risk of approaching a solution where decision boundaries cut through the unlabeled data increases. This is further amplified by the fact that a potentially large, high-dimensional feature space is likely to be sparsely populated, which allows for more incorrectly placed boundaries that satisfy the aim of entropy minimization algorithms. We tested our hypothesis by enabling CE during training and observed no improvement in results.

\subsection{Feature Normalization}
We found that standardizing the input feature vectors individually yields better results, as opposed to commonly used batch normalization \cite{batchnorm}, or no normalization at all. At first glance, it might seem counter-intuitive, as features hold sensitive semantics about their corresponding samples, and any type of normalization that does not take other statistics into consideration imposes the risk of harming the relationships between dimensions. However, this is done only as a pre-processing step, computing the standard score for vectors that are produced by the ResNet-101 model. Indeed, this is no different to widely popular image normalization \cite{dirt-t, temporalens}, as $G_r$ was not trained to output task-driven features. Therefore, we convert each input into zero-mean unit variance:
\begin{equation}
    \hat X_i = \frac{\hat X_i - \mu(\hat X_i)}{\sigma(\hat X_i)}
\end{equation}

\subsection{Iterative Refinement Training}
Decision-Boundary Iterative Refinement Training With a Teacher (DIRT-T) \cite{dirt-t} is a secondary training phase that disables the source signal in order to further minimize the error in the target domain. That leaves conditional entropy together with VAT, and introduces a consistency cost, to ensure that decision boundaries of a student model stay close to those of the teacher \cite{teacher}. Shu et al. \cite{dirt-t} note that DIRT-T suffers from high variance in results, and sometimes might lead to a degenerate solution, that is mainly caused by entropy minimization, which we discussed in the previous Subsection. Instead, we propose evaluating on an exponential moving average (EMA) of the network weights with decay 0.999. This idea is borrowed from the semi-supervised and generative fields of computer vision \cite{fixmatch, stylegan}, and is proven to consistently provide better results than the original (student) model, due to smooth weight updates. Although we see merit in disabling source signal and having a secondary training phase, that is rather a topic for future work than focus of this paper.

\section{Datasets}\label{Datasets}
\begin{figure}[!h]
  \includegraphics[width=\columnwidth]{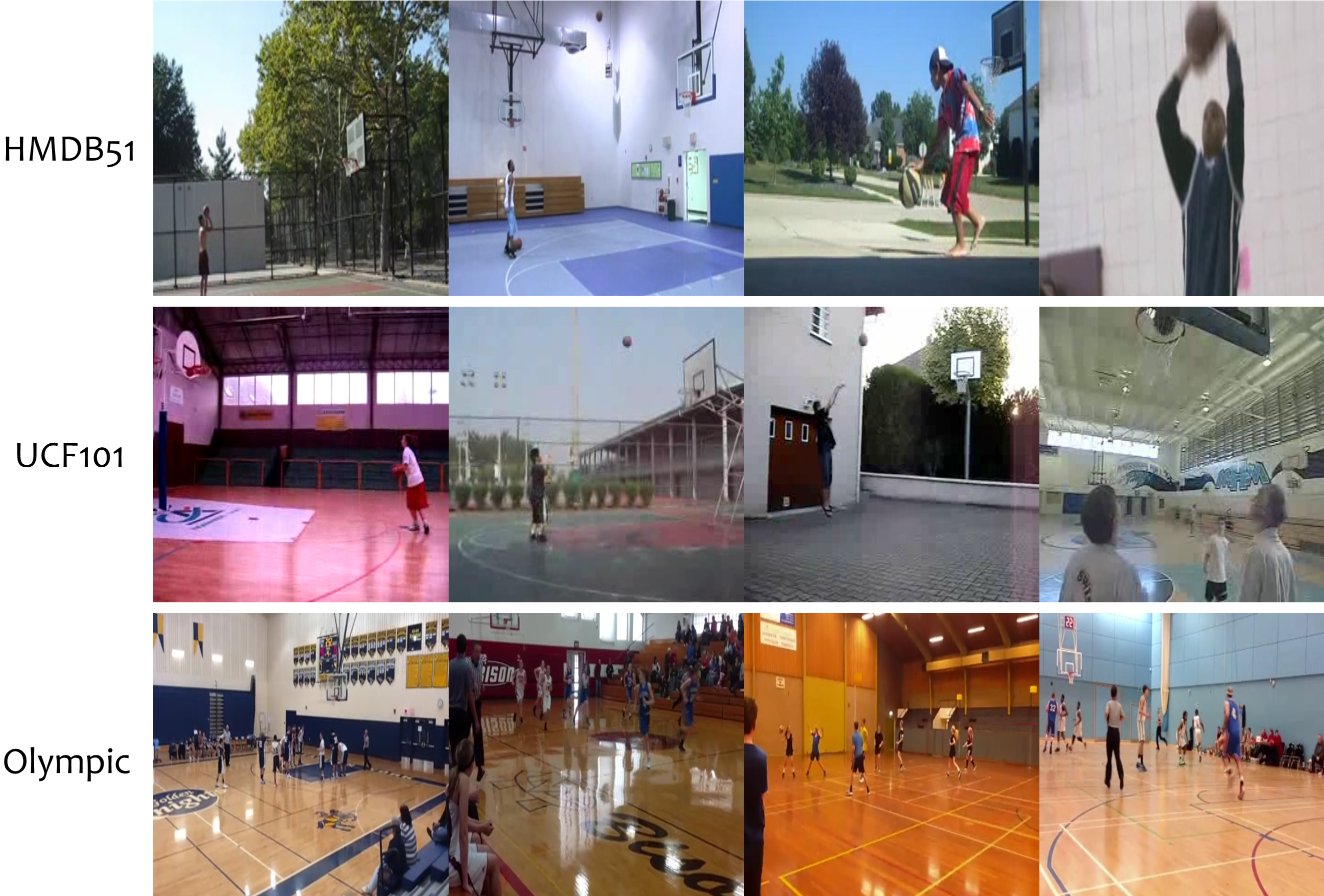}
  \caption{Snapshots of all three chosen datasets, from the overlapping category basketball/shoot ball.}
  \label{snapshots}
\end{figure}

The datasets chosen for evaluation are UCF101 \cite{ucf}, HMDB51 \cite{hmdb}, and Olympic Sports Dataset \cite{olympic}. All of them comprise of clips that were collected from public databases, such as YouTube, Google videos, and Prelinger archive. The videos show humans in motion, therefore, there is a significant overlap between categories. A detailed summary of adaptation paths formed by the datasets is presented in Table \ref{datasetstable}, while Fig. \ref{snapshots} shows snapshot examples of the basketball/shoot ball class. UCF\textsubscript{f}-HMDB\textsubscript{f} (f for ``full") is simply an extended version of UCF\textsubscript{s}-HMDB\textsubscript{s} (s for ``small"), which is designed to include extra samples and overlapping categories, making it a more challenging adaptation path than its predecessor \cite{ta3npaper}.
\begin{table}[!h]
\caption{Table 1: The summary of adaptation paths.}
\label{datasetstable}
\begin{center}       
\begin{tabular}{|c|c|c|c|} 
\hline
Path & Classes & Train videos & Test videos \\
\hline
UCF\textsubscript{s}-Olympic & 6 & 601-250 & 240-54 \\
UCF\textsubscript{s}-HMDB\textsubscript{s} & 5 & 482-350 & 189-150 \\
UCF\textsubscript{f}-HMDB\textsubscript{f} & 12 & 1438-840 & 571-360\\
\hline
\end{tabular}
\end{center}
\end{table} 

\section{Experiments}\label{Experiments}
The changes we made to the TA$^3$N implementation are as follows:
\begin{table*}[!h]
\caption{Table 2: The accuracy (\%) comparison with other methods on publicly available video UDA benchmark datasets.}
\label{results}
\begin{center}       
\begin{tabular}{|c|c|c|c|c|c|c|} 
\hline
Method & U\textsubscript{s}$\rightarrow$O & O$\rightarrow$U\textsubscript{s} & U\textsubscript{s}$\rightarrow$H\textsubscript{s} & H\textsubscript{s}$\rightarrow$U\textsubscript{s} & U\textsubscript{f}$\rightarrow$H\textsubscript{f} & H\textsubscript{f}$\rightarrow$U\textsubscript{f} \\
\hline
W. Sultani et al. \cite{sultani} & 33.33 & 47.91 & 68.70 & 68.67 & - & -\\
T. Xu et al. \cite{xu} & 87.00 & 75.00 & 82.00 & 82.00 & - & - \\
AMLS (GFK) \cite{jamal} & 84.65 & 86.44 & 89.53 & 95.36 & - & - \\
AMLS (SA) \cite{jamal} & 83.92 & 86.07 & 90.25 & 94.40 & - & - \\
DAAA \cite{jamal} & 91.60 & 89.96 & - & - & - & - \\
TA$^3$N \cite{ta3npaper} & 98.15 & 84.58 & 98.00 & \textbf{98.94} & 73.05 & 77.23 \\
\hline
Ours (TA$^3$N + VAT) & \textbf{100.0} & \textbf{90.00} & \textbf{98.67} & 98.42 & \textbf{79.73} & \textbf{84.07} \\
\hline
\end{tabular}
\end{center}
\end{table*} 
\begin{enumerate}
    \item Switched stochastic gradient descent (SGD) optimizer to Adam;
    \item Disabled learning rate and weight decay;
    \item Added VAT regularization;
    \item Added z-score normalization to the general-purpose features, $\hat X$;
    \item Performed evaluation on the EMA model.
\end{enumerate}
To keep the comparison with the initial implementation fair, we left all the hyperparameters and evaluation protocols the same \cite{ta3npaper}. We also use the same pre-extracted features, available at TA$^3$N repository\footnote{\url{https://github.com/cmhungsteve/TA3N}}. We note that results reported in the TA$^3$N paper differ from the ones available at the repository, therefore, we report the latter, as that is the code we worked with (Table \ref{results}). For the other approaches, we drew the results from their original papers. We evaluate our system on 6 adaptation paths:
\begin{itemize}
    \item \textbf{UCF\textsubscript{s} - Olympic}, denoted by U\textsubscript{s}$\rightarrow$O and O$\rightarrow$U\textsubscript{s};
    \item \textbf{UCF\textsubscript{s} - HMDB\textsubscript{s}}, denoted by U\textsubscript{s}$\rightarrow$H\textsubscript{s} and H\textsubscript{s}$\rightarrow$U\textsubscript{s};
    \item \textbf{UCF\textsubscript{f} - HMDB\textsubscript{f}}, denoted by U\textsubscript{f}$\rightarrow$H\textsubscript{f} and H\textsubscript{f}$\rightarrow$U\textsubscript{f};
\end{itemize}
For 5 out of 6 the paths, the suggested changes improve upon the current state of the art, reducing the error by up to 6.84\%. For H\textsubscript{s}$\rightarrow$U\textsubscript{s}, they maintain competitive performance. The results are available in Table \ref{results}.

\section{Conclusions and Future Work}\label{Conclusion and Future Work}
In this paper, we proposed several approaches for unsupervised video domain adaptation: Virtual Adversarial Training as well as normalization of input features, and testing on the exponential moving average weights. We empirically show that when combined together, these techniques significantly improve the performance of the recent state of the art model called TA$^3$N. The proposed changes are independent of each other, easy to implement, applicable to other deep learning tasks, do not add much computational overhead, and can be combined with most existing neural network architectures. We stress that performing VAT on features is faster than performing VAT on individual frames, as it does not interfere with the feature extraction parts of the system. This property opens up additional opportunities for methods that work with pre-extracted features and/or  have multiple networks, such as encoder/decoder models.

For future work, we would like to do the following:
\begin{enumerate}
    \item Apply VAT in feature space to other deep learning problems;
    \item Explore augmentation methods in the temporal domain to enhance video classification and DA;
    \item Create new, more challenging adaptation paths for video DA;
    \item Stabilize DIRT-T and make it less dependent on the teacher model.
\end{enumerate}

\section{Acknowledgments}
The authors are grateful for the support from the Natural Environment Research Council and Engineering and Physical Sciences Research Council through the NEXUSS Centre for Doctoral Training (grant \#NE/RO12156/1).


\begin{biography}
Artjoms Gorpincenko received the bachelor's degree in computing science from the University of East Anglia, Norwich, U.K., in 2018. He is currently pursuing the Ph.D. degree at the University of East Anglia. His research interests are related to computer vision and deep learning.

Geoffrey French received the bachelor's and master's degrees in computing science from the University of East Anglia, Norwich, U.K., in 2001 and 2013, respectively. He is currently pursuing the Ph.D. degree with the University of East Anglia. His research interests include computer vision and deep learning.

Michal Mackiewicz is an Associate Professor at the School of Computing Sciences, University of East Anglia (UEA). He received his MSc from the University of Science and Technology (AGH), Krakow, Poland in 2003 and the PhD from UEA in 2008. Michal has been involved in researching areas of colour science, physics based vision and machine (and deep) learning. He has worked on a number of computer vision applications including medical imaging, remote sensing, environmental monitoring and agri-tech.

\end{biography}

\end{document}